\documentclass[letterpaper, 10 pt, conference]{ieeeconf}  

\IEEEoverridecommandlockouts                              

\usepackage{graphics}
\usepackage{amsmath}
\usepackage{amssymb}
\usepackage{graphicx}
\usepackage{colortbl}     
\usepackage{xcolor}       
\usepackage{booktabs} 
\usepackage{subcaption}
\usepackage{hyperref}
\hypersetup{
    colorlinks=true,
    linkcolor=blue!60!black,
    citecolor=blue!60!black,
    urlcolor=blue!60!black
}
\usepackage{cleveref}
\crefname{figure}{Fig.}{Figs.}
\crefname{table}{TABLE}{TABLES}
\usepackage{epsfig}
\usepackage{mathptmx}
\usepackage{times}
\usepackage{cite}

\title{\LARGE \bf
DRCP: Diffusion on Reinforced Cooperative Perception for Perceiving Beyond Limits
}

\author{Lantao Li$^{1}$, Kang Yang$^{1,2}$, Rui Song$^{3}$ and Chen Sun$^{1}$ 
\thanks{    
        {\tt\small   }}%
\thanks{    
        {\tt\small   }}%
\thanks{$^{1}$Sony (China) Limited
        {\tt\small lantao.li@sony.com}}%
\thanks{$^{2}$Renmin University of China
        }%
\thanks{$^{3}$Fraunhofer IVI
        }%
}

\begin{document}

\maketitle
\thispagestyle{empty}
\pagestyle{empty}

\begin{abstract}

Cooperative perception enabled by Vehicle-to-Everything communication has shown great promise in enhancing situational awareness for autonomous vehicles and other mobile robotic platforms. Despite recent advances in perception backbones and multi-agent fusion, real-world deployments remain challenged by hard detection cases, exemplified by partial detections and noise accumulation which limit downstream detection accuracy. This work presents Diffusion on Reinforced Cooperative Perception (DRCP), a real-time deployable framework designed to address aforementioned issues in dynamic driving environments. DRCP integrates two key components: (1) Precise-Pyramid-Cross-Modality-Cross-Agent, a cross-modal cooperative perception module that leverages camera-intrinsic-aware angular partitioning for attention-based fusion and adaptive convolution to better exploit external features; and (2) Mask-Diffusion-Mask-Aggregation, a novel lightweight diffusion-based refinement module that encourages robustness against feature perturbations and aligns bird's-eye-view features closer to the task-optimal manifold. The proposed system achieves real-time performance on mobile platforms while significantly improving robustness under challenging conditions. Code will be released in late 2025.

\end{abstract}

\section{INTRODUCTION}

Robotic systems such as autonomous vehicles and mobile agents rely heavily on perception to understand their surroundings and make informed decisions. Nevertheless, single-agent perception faces critical limitations in real-world environments, where occlusions in crowded traffic and degraded sensor performance under adverse lighting often lead to incomplete or inaccurate environmental understanding.

To address these challenges, cooperative perception has emerged as a promising paradigm, enabling multiple agents to share sensory data or intermediate features via vehicle-to-everything (V2X) communication. By aggregating observations from different viewpoints, cooperative systems can improve robustness and coverage, particularly in regions occluded to single-agent perception. However, achieving reliable and efficient multi-agent, multi-modal fusion remains difficult. Recent bird's-eye-view (BEV) based methods such as HEAL \cite{c2} and CoBEVT \cite{c3}, for example, achieve their best performances in LiDAR-only settings, while their accuracy degrades in LiDAR–camera configurations. This degradation stems from camera-track BEV generation (e.g., LSS \cite{lss} backbones), where explicit depth estimation is used to construct camera BEV but undermines LiDAR BEV cues during cross-modal BEV-to-BEV fusion. Consequently, fully leveraging the complementary strengths of multi-modal inputs remains challenging. Furthermore, fusing features across agents into a coherent representation requires further exploration. Developing more effective cross-modal backbones and integrated cross-agent fusion frameworks is thus crucial.

\begin{figure}[t]
  \centering
   \includegraphics[width=1.0\linewidth]{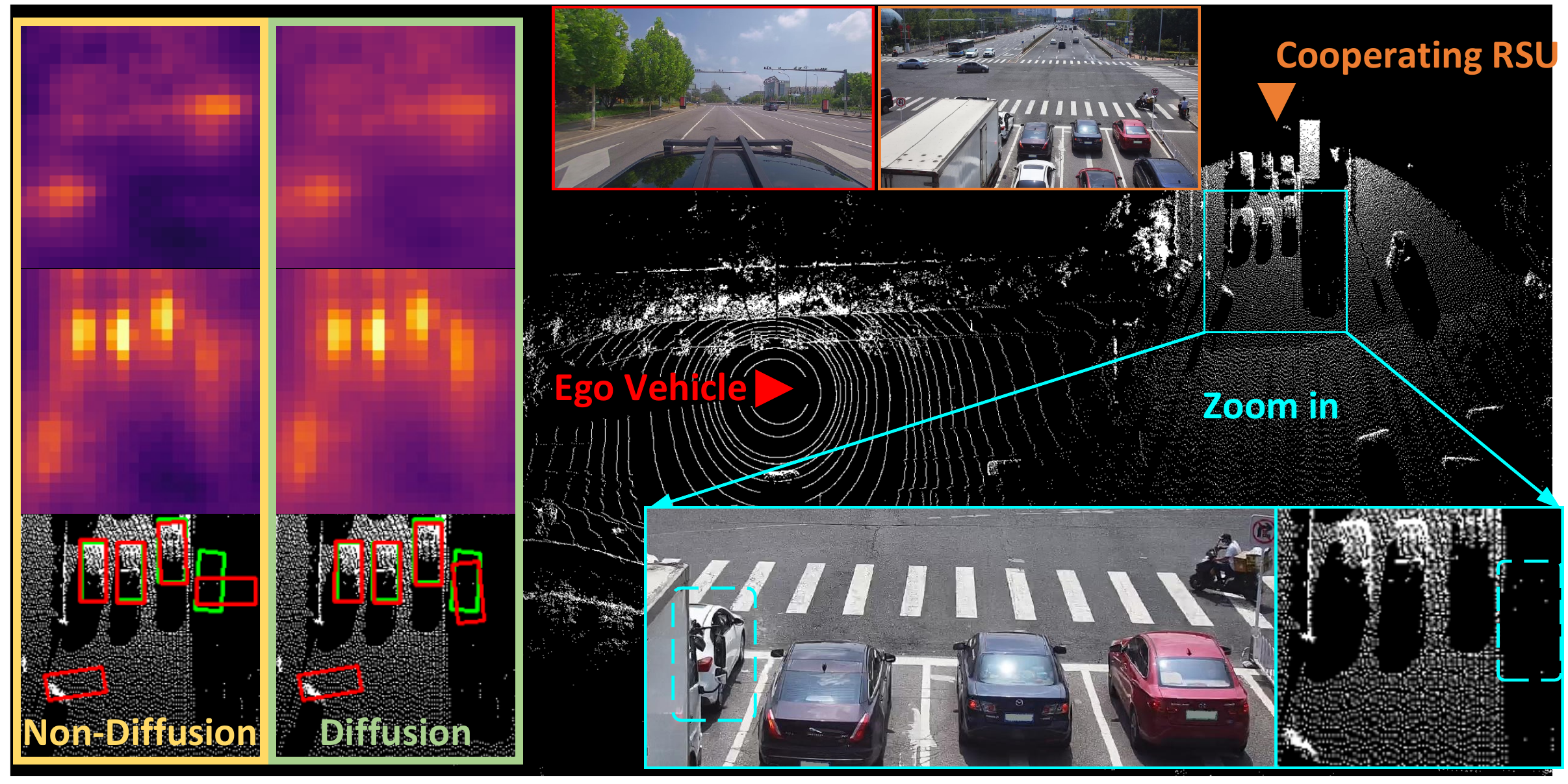}
   \caption{Heatmaps of the classification head show that in the non-diffusion setting (left), horizontal responses (upper row) dominate over vertical ones (middle row), leading to 90° misalignment in bounding box prediction (bottom row). With diffusion (right), horizontal noise is suppressed while vertical cues are strengthened, correcting orientation via denoising ambiguous BEV features.}
   \label{fig:concept}
\end{figure}

Real-world sensory inputs and conventional backbone processing often produce BEV features that deviate from task-optimal manifolds, resulting in suboptimal representations for downstream perception. Inspired by diffusion models, which iteratively refine noisy data, we propose a lightweight, single-step diffusion module to adaptively enhance BEV features. Specifically, the module generates a channel-wise conditioning seed to guide a single-step denoising of deliberately perturbed BEV features, producing complementary residuals that are selectively fused with the original BEV. This process amplifies or attenuates ambiguous features according to learned patterns, aligning the final BEV representation with the task-optimal manifold. Despite its simplicity, this strategy consistently improves multi-agent BEV perception, as illustrated in \cref{fig:concept}.

To this end, we present \textbf{Diffusion on Reinforced Cooperative Perception (DRCP)}, a real-time cooperative perception framework that integrates two key components:
\begin{itemize}
    \item \textbf{Precise Pyramid Cross-Modal Cross-Agent (PPXX)} fusion module that projects multi-camera features onto LiDAR BEV maps using camera intrinsics, followed by adaptive multi-scale integration across agents.
    \item \textbf{Mask-Diffusion-Mask-Aggregation (MDMA)}, a diffusion-based refinement in BEV space that denoises features and aligns them to the task-optimal manifold.
\end{itemize}

The overall framework is optimized for real-time deployment and achieves state-of-the-art performance on major cooperative perception benchmarks (DAIR-V2X \cite{dair} and OPV2V \cite{opv2v}), demonstrating both effectiveness and efficiency.

\section{Related Works}

\subsection{Cooperative Perception}
Single-agent perception has advanced with camera depth estimation (LSS \cite{lss}), LiDAR voxel encoding (VoxelNet \cite{VoxelNet}), and transformer-based attention \cite{vit, Swin}. BEV-based fusion \cite{BEVFormer, las} unifies feature space, while occupancy networks \cite{occ, pyocc} provide continuous scene representations. Cooperative perception extends these via multi-agent collaboration, supported by datasets \cite{dair, opv2v, v2v4real, tum, v2xradar}. Fusion evolved from raw \cite{early} and late \cite{late} strategies to intermediate feature fusion with transformers \cite{c0, c1, c2, bm2cp, c3}, balancing efficiency and accuracy. Adaptive schemes (Who2Com, Where2Com \cite{c4, c5}) dynamically select agents/regions to save bandwidth; V2X works \cite{v2x1, v2x2} optimize wireless sharing; FedBEVT \cite{fedbevt} explores decentralized training. Alignment methods (CoAlign, CBM \cite{coalign, cbm}) mitigate localization errors, and vision-action systems \cite{coopernaut, icop} show end-to-end benefits. Recent studies target multi-modal cooperation: HM-ViT \cite{c1} and HEAL \cite{c2} adopt heterogeneous single-modality agents, excelling in LiDAR-only cases, while BM2CP \cite{bm2cp} allows multi-modal inputs but still lags behind HEAL, showing the difficulty of exploiting cross-modal complementarity.  

\subsection{Diffusion Models}
Diffusion models, successful in generation, are emerging in perception \cite{difbox, difbev, diffuser}. Diff3Det \cite{difbox} treats detection as denoising, DiffBEV \cite{difbev} refines BEV features, and DifFuser \cite{diffuser} fuses multi-modal BEV via gated self-conditioning. One-step variants \cite{onestep1, onestep2} achieve high-quality results with a single denoising step, enabling fast super-resolution and anomaly detection, motivating lightweight one-step refinement in BEV. For cooperation, diffusion has been used to reconstruct BEV from compressed semantics \cite{difcp} (saving bandwidth but losing precision) and to integrate radar-conditioned LiDAR features \cite{v2xr} (enhancing robustness). Yet no prior work shows diffusion boosting cooperative detection in real time with real-world data, motivating our design.

\section{Methodology}

\subsection{PPXX for evolved fusion}
\label{sec:ppxx}

\noindent \textbf{Camera-Intrinsics-Aware Radian Division:} 
Our cross-modal fusion design enhances LiDAR features with the complementary semantic richness of camera inputs while avoiding explicit depth estimation from cameras to preserve LiDAR depth accuracy. Building on prior work \cite{las} that explored perspective-to-BEV projection for single-agent fusion, we address its limitation of ignoring camera intrinsics in horizontal angular divisions. We propose Intrinsics-Radian-Glue Attention (Inrin-RG-Attn), which refines the projection by ``gluing" LiDAR and camera features based on radian correspondence derived from camera intrinsics calibration, leveraging a polar projection that balances near- and far-field sampling and preserves angular continuity.

\begin{figure*}[!ht]
  \vspace{2mm}
  \centering
   \includegraphics[width=1.0\linewidth]{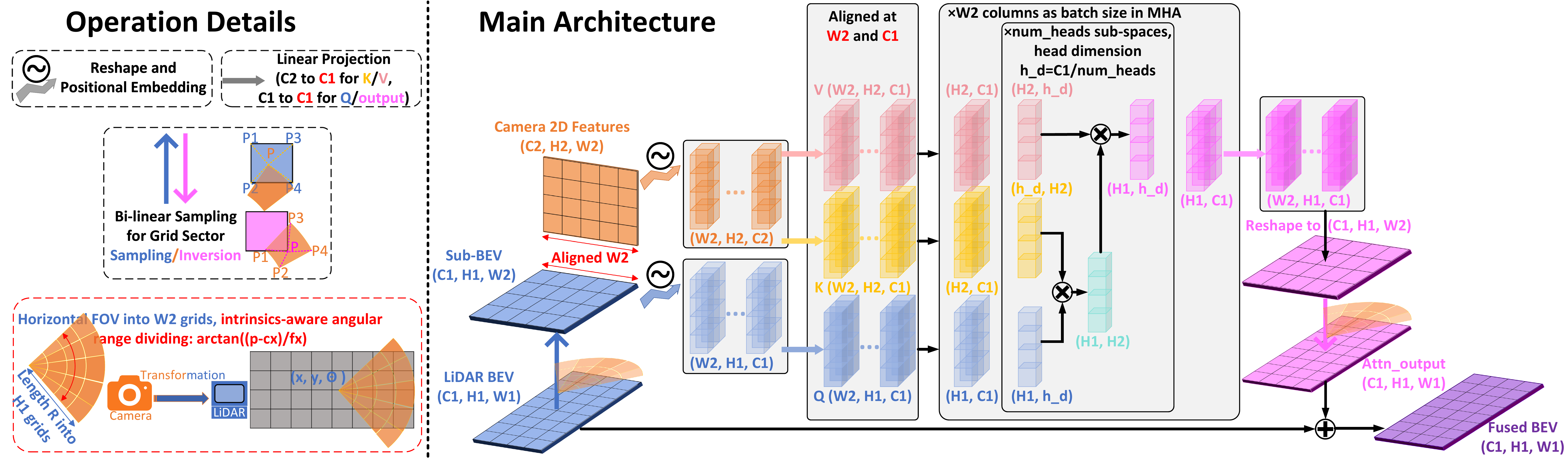}
   \caption{Workflow of Camera-Intrinsics-Aware Radian Division for column-to-column Radian-Glue Attention. The calibrated radian division establishes accurate column-wise correspondence between modalities, enabling efficient attention-based cross-modal fusion and feature enhancement.}
   \label{fig:intrin}
\end{figure*}

For an agent equipped with LiDAR and cameras, let the BEV feature map from LiDAR be $F^{BEV} \in \mathbb{R}^{C_1 \times H_1 \times W_1}$, and the 2D semantic feature map from the $i{\text{-th}}$ camera be $F^{i{\_cam}} \in \mathbb{R}^{C_2 \times H_2 \times W_2}$. The extrinsic transformation $\mathbf{T}^{BEV}_{i{\_cam}} = [\mathbf{R}^{BEV}_{i{\_cam}} | \mathbf{t}^{BEV}_{i{\_cam}}]$ aligns two coordinate systems, where $\mathbf{R}^{BEV}_{i{\_cam}} \in \mathbb{R}^{3 \times 3}$ is the rotation matrix and $\mathbf{t}^{BEV}_{i{\_cam}} \in \mathbb{R}^{3}$ is the translation vector. The camera position and the orientation of its optical axis in the BEV frame are obtained as:

\begin{equation}
    \mathbf{p}^{BEV}_{i{\_cam}} = \mathbf{R}^{BEV}_{i{\_cam}} \mathbf{p}_{i{\_cam}} + \mathbf{t}^{BEV}_{i{\_cam}},
    \mathbf{d}^{BEV}_{i{\_cam}} = \mathbf{R}^{BEV}_{i{\_cam}} 
    \begin{bmatrix}
        \cos\theta_{i{\_cam}} \\
        \sin\theta_{i{\_cam}} \\
        0
    \end{bmatrix},
\end{equation}
where $\mathbf{p}_{i{\_cam}}$ is the camera position in its own frame, and $\theta_{i{\_cam}}$ is the horizontal orientation angle of the optical axis measured from the local $x$-axis. The transformed optical center direction in the BEV frame is:
\begin{equation}
    \theta^{BEV}_{i{\_cam}} = \arctan \left( \frac{d^{BEV}_{i{\_cam}, y}}{d^{BEV}_{i{\_cam}, x}} \right).
\end{equation}
This defines the camera’s horizontal field of view (FOV) in the BEV frame, which is discretized into $W_2$ angular sub-sectors corresponding to the columns of $F^{i{\_cam}}$. Each column corresponds to a width on the original sensor:
\begin{equation}
p_{\text{width}} = \frac{W_{\text{camera resolution}}}{W_2}, \quad
p_{\text{origin}, m} = m \cdot p_{\text{width}} + \frac{p_{\text{width}}}{2},
\end{equation}
where $m$ denotes the $m{\text{-th}}$ column. Relative offsets from the principal point $c_x$ are:
\begin{equation}
\Delta p_m = \frac{p_{\text{origin}, m} - c_x}{f_x},
\end{equation}
where $f_x$ is the focal length in pixels. The angular direction in the BEV frame for each column is:
\begin{equation}
\theta_m = \arctan(\Delta p_m) + \theta^{BEV}_{i{\_cam}}.
\end{equation}
To extract the corresponding sub-region from the BEV map, we sample radial distances uniformly:
\begin{equation}
    r_n = \frac{n}{H_1} R, \quad R = \frac{W_1}{2}, \quad n = 1, 2, \dots, H_1.
\end{equation}
The sampling coordinates in the BEV map are:
\begin{equation}
    x_{m, n} = {p}^{BEV}_{i{\_cam}, x} + r_n \cos\theta_m,
    y_{m, n} = {p}^{BEV}_{i{\_cam}, y} - r_n \sin\theta_m.
\end{equation}

Bilinear interpolation, employed as Grid Sector Sampling, extracts a sub-BEV map and yields a rectangular representation $F^{sub\_BEV} \in \mathbb{R}^{C_1 \times H_1 \times W_2}$. After reshaping, positional embedding, and a channel-alignment linear projection, the features (aligned along width $W_2$ and channels $C_1$) are fused via multi-head attention (MHA) in a column-to-column manner: LiDAR BEV features serve as the Query, while camera features serve as the Key and Value. This column-wise attention treats $W_2$ as the batch dimension of MHA, thereby enabling efficient computation. The fused features are then projected back through inverse bilinear sampling and combined with the original LiDAR BEV via element-wise addition, thereby integrating camera-enhanced semantics. The overall camera-intrinsics-aware fusion process is detailed in \cref{fig:intrin} and formulated as:
\begin{equation}
{F^{Fused\_BEV}} = \text{Intrin-RG-Attn}(F^{BEV},\ F^{i{\_cam}}).
\end{equation}
The procedure can be applied independently for multiple onboard cameras.

\begin{figure*}[!ht]
  \vspace{2mm}
  \centering
   \includegraphics[width=1.0\linewidth]{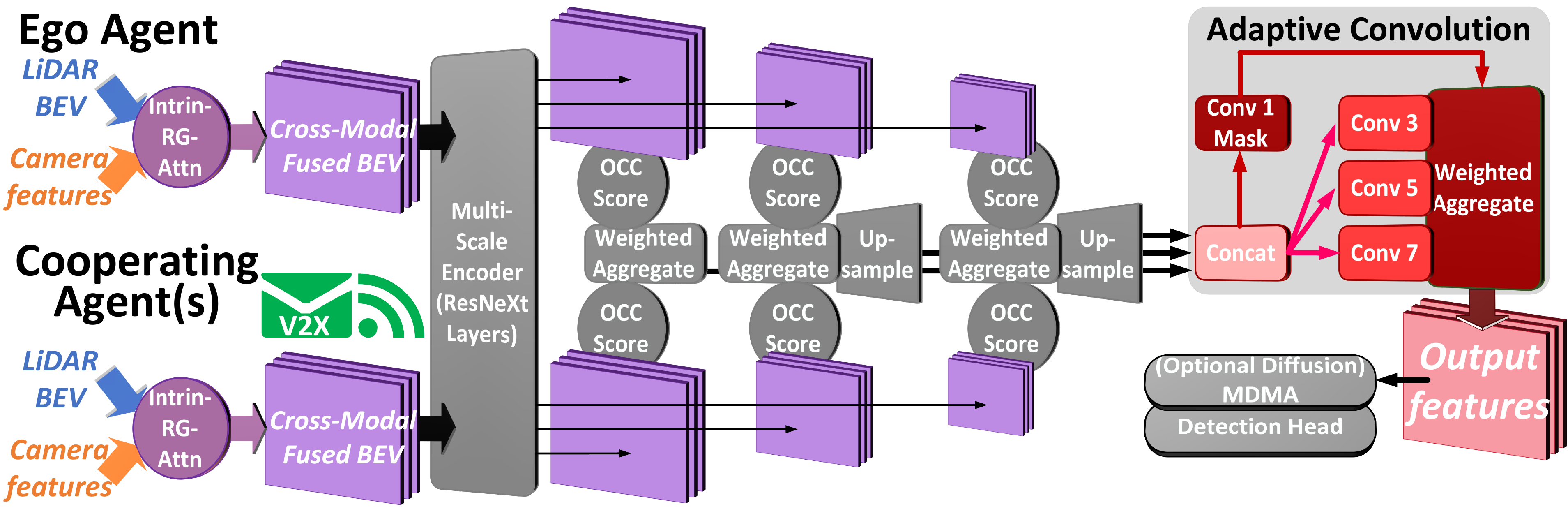}
   \caption{The overall structure of PPXX module (\cref{sec:ppxx}) for conducting fusion across-modal (Intrin-RG-Attn) and across-agent (Adaptive Convolution) in the fully integrated pyramid manner.}
   \label{fig:ppxx}
\end{figure*}

\noindent \textbf{Integrated Pyramid Fusion:}
After cross-modal fusion via Intrin-RG-Attn on each agent’s high-resolution LiDAR and camera features, we construct a hierarchical multi-scale representation to support downstream reasoning (see \cref{fig:ppxx}). Specifically, we define
\begin{equation}
    \{\,F^{{Fused\_BEV}}_{k,(s)}\in\mathbb{R}^{C_s\times H_s\times W_s}\}_{s=1}^{3},
\end{equation}
where \(F^{{Fused\_BEV}}_{k,(s)}\) is the fused BEV feature of agent \(k\) at scale \(s\), covering the original resolution and two progressively coarser scales (e.g., \(W=256,128,64\)).

To guide cross-agent fusion, each scale produces a cell-wise occupancy score map \(\text{occ}_{k,(s)}\in\mathbb{R}^{H_s\times W_s}\) via a shared convolutional head. During training, these scores are supervised by binary BEV occupancy labels derived from ground-truth object classification annotations: cells overlapping any object are labeled as occupied (positive), and all others as free (negative), using a sigmoid focal loss. BEV features from other agents are transformed into the ego frame via V2X, and we compute per-agent weights
\begin{equation}
    \alpha_{k,(s)}
    = \frac{\text{occ}_{k,(s)}}{\sum_{l=1}^{N}\text{occ}_{l,(s)}}, \quad k=1,\dots,N,
\end{equation}
where \(N\) is the number of agents. We fuse via
\begin{equation}
    F^{{Fused\_BEV}}_{(s)}
    = u_s \sum_{k=1}^{N}\alpha_{k,(s)}\,F^{{Fused\_BEV}}_{k,(s)},
\end{equation}
with \(u_s\) upsampling to a unified scale. Finally, we concatenate across scales:

\begin{equation}
  F^{{Fused\_BEV}}_{pyr}=\mathop{\mathrm{Concat}}_s\bigl(F^{{Fused\_BEV}}_{(s)}\bigr).
\end{equation}

\noindent \textbf{Adaptive Convolution at Final BEV:}
Although the integrated pyramid fusion module aggregates cross-modal features from all collaborative agents, its occupancy-based aggregation does not model inter-agent feature interactions. To further refine the final BEV representation, we attach a dynamic multi-scale convolution fusion module, which adaptively calibrates features corresponding to the same object across different scales as shown in \cref{fig:ppxx}.

Given the fused BEV feature map \(F^{{Fused\_BEV}}_{pyr} \in \mathbb{R}^{C \times H \times W}\), we perform three convolution operations with kernel sizes \(3\times3\), \(5\times5\), and \(7\times7\), respectively:
\begin{equation}
\begin{split}
    f_3 &= \text{Conv}_3\bigl(F^{{Fused\_BEV}}_{pyr}\bigr),\\[1mm]
    f_5 &= \text{Conv}_5\bigl(F^{{Fused\_BEV}}_{pyr}\bigr),\\[1mm]
    f_7 &= \text{Conv}_7\bigl(F^{{Fused\_BEV}}_{pyr}\bigr).
\end{split}
\end{equation}
where \(f_3, f_5, f_7 \in \mathbb{R}^{C \times H \times W}\). Simultaneously, a dynamic weight generator (a \(1\times1\) convolution followed by softmax) produces a weight tensor $w \in \mathbb{R}^{3 \times H \times W}$, ensuring that for every spatial location \((h, w)\):
\begin{equation}
    w_1(h,w) + w_2(h,w) + w_3(h,w) = 1.
\end{equation}

The final refined BEV feature is computed as:
\begin{equation}
    F^{{Fused\_BEV}}_{final} = w_1 \odot f_3 + w_2 \odot f_5 + w_3 \odot f_7,
\end{equation}
where \(\odot\) denotes element-wise multiplication.

\subsection{MDMA for Diffusion on BEV}

\begin{figure}[t]
  \centering
   \includegraphics[width=1.0\linewidth]{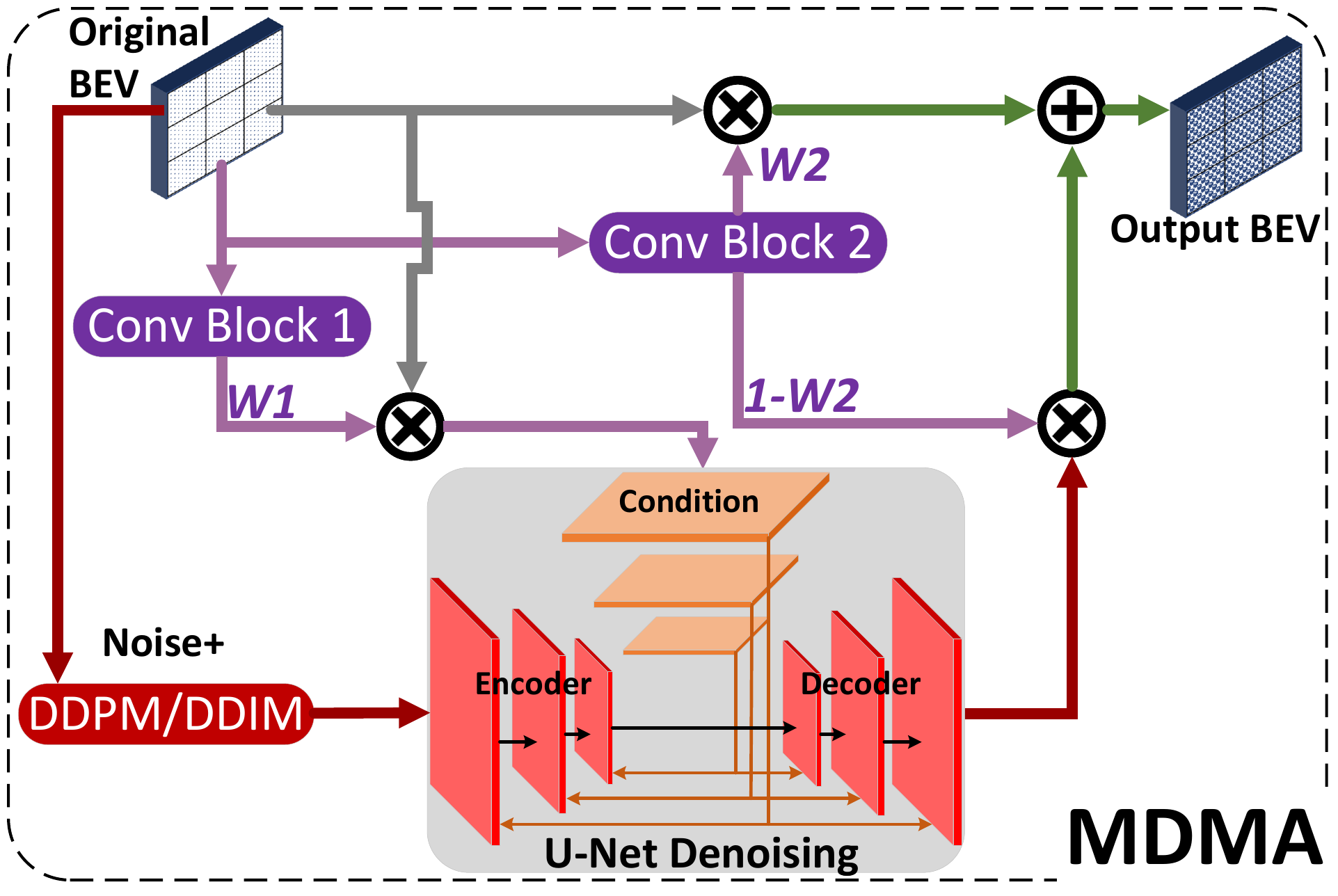}
   \caption{Overview of the MDMA module. A channel-wise seed extracts reliable channels to condition a single-step diffusion denoiser; a learned residual mask then adaptively fuses denoised corrections with the original BEV to yield an enhanced representation.}
   \label{fig:mdma}
\end{figure}

The module in \cref{fig:mdma} implements a lightweight, task-oriented refinement for BEV feature maps. Given an input
\begin{equation}
  F^{BEV}_{origin} \in \mathbb{R}^{C\times H\times W},
\end{equation}
we adopt a diffusion-based workflow—seed condition extraction, forward perturbation, single-step conditioned denoising, and residual fusion—designed explicitly as regularization for downstream perception heads.

\noindent\textbf{Seed condition extraction:} A channel-wise confidence mask is predicted by a $1\times1$ convolution and sigmoid:
\begin{equation}
  W_1=\sigma\bigl(\mathrm{conv}_1(F^{BEV}_{origin})\bigr),
\end{equation}
which produces the conditioned seed as:
\begin{equation}
  F^{BEV}_{seed}=F^{BEV}_{origin}\odot W_1.
\end{equation}
The seed condition (channel-wise scaled original features) highlights reliable components and constrains the conditioning magnitude to keep the subsequent one-step denoising numerically stable.

\noindent\textbf{Forward perturbation:} Conceptually, the forward process is modeled as Gaussian corruption:
\begin{equation}
  F^{BEV}_{dif,t}=\sqrt{\bar{\alpha}_t}\,F^{BEV}_{origin}+\sqrt{1-\bar{\alpha}_t}\,\boldsymbol{\epsilon},\qquad\boldsymbol{\epsilon}\sim\mathcal{N}(0,I),
\end{equation}
which provides a structured noisy input for refinement, without intermediate reconstruction losses.

\noindent\textbf{single-step conditioned denoising:} The reverse trajectory is collapsed into a deterministic, seed-conditioned mapping:
\begin{equation}
\hat{F}^{BEV}_{clean}=\mathcal{D}_{\theta}\bigl(F^{BEV}_{dif,t},\,t,\,F^{BEV}_{seed}\bigr),
\end{equation}
where $\mathcal{D}_{\theta}$ is implemented as a compact U-Net with two down-sampling and two up-sampling blocks. The mapping is treated as a single-step deterministic refinement (DDIM-like collapse).

\noindent\textbf{Residual fusion:} A second $1\times1$ convolution predicts an adaptive interpolation mask:
\begin{equation}
  W_2=\sigma\bigl(\mathrm{conv}_2(F^{BEV}_{origin})\bigr).
\end{equation}
used to combine original and denoised features:
\begin{equation}
  F^{BEV}_{final} = F^{BEV}_{origin}\odot W_2 \;+\; \hat{F}^{BEV}_{clean}\odot(1-W_2),
\end{equation}
or equivalently:
\begin{equation}
  F^{BEV}_{final} = F^{BEV}_{origin} \;+\; \Bigl(\hat{F}^{BEV}_{clean}-F^{BEV}_{origin}\Bigr)\odot(1-W_2),
\end{equation}
where the residual term acts as a data-adaptive posterior correction: it injects small, stable compensatory signals only where beneficial.

\noindent\textbf{Training objective:} No intermediate loss is imposed for exact reconstruction of $F^{BEV}_{origin}$ (e.g., on $\hat{F}^{BEV}_{clean}$). Instead, the module is trained end-to-end with downstream task losses (classification, regression, direction), enabling the denoiser and masks to learn task-optimal and numerically stable refinements rather than literal reconstructions.

\section{Experiments}

\subsection{Datasets \& Settings}
\noindent \textbf{Datasets:} We evaluate our approach on two datasets: DAIR-V2X \cite{dair} and OPV2V \cite{opv2v}. DAIR-V2X is a real-world dataset from Beijing's Autonomous Driving Zone comprising 9K frames, each containing LiDAR and 1920×1080 camera data from both a vehicle and an RSU. Notably, the RSU’s LiDAR has 300 channels with a 100° FOV, while the vehicle’s LiDAR offers 40 channels over a 360° FOV. In contrast, OPV2V is a CARLA-based simulated dataset with over 11K frames covering diverse scenarios, featuring 2–7 vehicles per frame. Each vehicle is equipped with a 64-channel 360° LiDAR and four 800×600 cameras.

\noindent  \textbf{Settings:} Our architecture builds upon HEAL \cite{c2} by incorporating the MDMA diffusion module (as a final attachment) and replacing the BEV-related modules with our PPXX module. For fair comparison, raw data processing remains consistent: LiDAR point clouds are encoded using PointPillar, and camera images are processed with the first five layers of ResNet101. For pyramid cross-agent fusion, features are configured to fuse at widths of 64, 128, and 256 respectively. The detection range is set to \(x\in[-102.4,102.4]\) m and \(y\in[-51.2,51.2]\) m, and average precision (AP) is computed at various IoU thresholds (e.g., AP30 for AP@IoU=0.3).

\subsection{Training \& Inference Details}

Given the final BEV features of size $(256,128,256)$, we adopt an anchor-based detection design with $N_{\text{anchor}}=6$ anchors per spatial location. The detection heads are defined as: a classification head $\mathrm{Conv}(256,6,1)$, a regression head $\mathrm{Conv}(256,7 \times 6,1)$ with 3D box parameterization $(x,y,z,h,w,l,\theta)$, a direction head $\mathrm{Conv}(256,2 \times 6,1)$ modeling orientation via 2-bin classification, and an occupancy head $\mathrm{Conv}(256,1,1)$. During inference, final bounding boxes are obtained by converting classification logits to probabilities, combining regression and orientation outputs, and applying non-maximum suppression.

The PPXX-only non-diffusion architecture is trained end-to-end, whereas the MDMA module requires a two-stage scheme: a non-diffusion baseline is first optimized, then fine-tuned with the diffusion module for BEV enhancement, skipping standard multi-step DDPM/DDIM and performing a single-step, noise-driven refinement. The overall training objective is
\begin{align}
  \mathcal{L}_{\text{total}} = &\, \lambda_{\text{reg}}\,\mathcal{L}_{\text{reg}}(F, y_{\text{reg}}) + \lambda_{\text{cls}}\,\mathcal{L}_{\text{cls}}(F, y_{\text{cls}}) \notag \\
  &\, + \lambda_{\text{dir}}\,\mathcal{L}_{\text{dir}}(F, y_{\text{dir}}) + \lambda_{\text{occ}}\,\mathcal{L}_{\text{occ}}(F, y_{\text{occ}}),
\end{align}
where \(\mathcal{L}_{\text{cls}}\) and \(\mathcal{L}_{\text{occ}}\) are both sigmoid focal losses 
(\(\alpha{=}0.25, \gamma{=}2.0, \lambda_{\text{cls}}{=}\lambda_{\text{occ}}{=}1.0\)), 
\(\mathcal{L}_{\text{reg}}\) is the weighted smooth L1 loss (\(\sigma{=}3.0, \lambda_{\text{reg}}{=}2.0\)), and \(\mathcal{L}_{\text{dir}}\) is the softmax cross-entropy loss for orientation (\(\lambda_{\text{dir}}{=}0.4\)).

For DAIR-V2X, we use two agents (the maximum available), while for OPV2V we adopt dynamic participation with 2–5 agents. The Adam optimizer is employed with an initial learning rate of 0.002 for non-diffusion training, decayed by a factor of 0.1 at epoch 27 for DAIR-V2X and at epoch 38 for OPV2V. For MDMA fine-tuning, the learning rate is set to 0.0001 with up to 3 training epochs. All training is performed on a single NVIDIA RTX 6000 Ada, and inference is conducted on an NVIDIA RTX 3060.

\subsection{Quantitative \& Visualization Results}

\noindent \textbf{Performance Comparison:} As shown in \cref{tab:perf}, our proposed DRCP (PPXX+MDMA) surpasses previous state-of-the-art methods by at least 4.6\%, 4.6\%, and 5.1\% in AP30, AP50, and AP70 on DAIR-V2X, a dataset that inherently contains real-world sensor noise and localization errors, providing a challenging benchmark to highlight substantial gains in detection accuracy and robustness. On OPV2V, DRCP also sets new state-of-the-art results, with improvements of 1.0\%, 1.3\%, and 2.5\% in AP30, AP50, and AP70, respectively. The comparatively smaller gains on OPV2V can be attributed to the cleaner simulated data with fewer object types and less localization noise, as well as the larger number of cooperating agents (up to five), which naturally mitigates partial detections and occlusions, reducing the potential for further improvement. Moreover, in a LiDAR-camera multi-modal configuration, methods such as HEAL and CoBEVT, which rely on camera-track BEV with explicit depth estimation, experience performance drops---DAIR-V2X AP30 0.588 (-19.9\%)/0.776 (-5.6\%) and OPV2V AP50 0.643 (-29.2\%)/0.854 (-10.9\%)---highlighting the detrimental effect of introducing noisy depth predictions from camera features.

\begin{table}[h]
\centering
\resizebox{\columnwidth}{!}{%
\begin{tabular}{c|c|c|c|c|c|c} 
\hline
Dataset & \multicolumn{3}{c|}{DAIR-V2X} & \multicolumn{3}{c}{OPV2V} \\
\hline
Method & AP30 & AP50 & AP70 & AP30 & AP50 &AP70\\
\hline
F-Cooper (L) \cite{fcooper} & 0.723 & 0.620 & 0.445 & 0.876 & 0.855 & 0.678\\

DiscoNet (L) \cite{c0} & 0.746 & 0.685 & 0.516 & 0.889 & 0.881 & 0.737\\

AttFusion (L) \cite{opv2v} & 0.713 & 0.644 & 0.511 & 0.875 & 0.859 & 0.749\\

V2XViT (L) \cite{v2xvit} & 0.785 & 0.724 & 0.553 & 0.952 & 0.934 & 0.854\\

CoBEVT (L) \cite{c3} & 0.787 & 0.692 & 0.532 & 0.943 & 0.935 & 0.851\\
 
HM-ViT (L) \cite{c1} & 0.818 & 0.761 & 0.601 & 0.956 & 0.950 & 0.873\\

BM2CP (LC) \cite{bm2cp} & 0.802 & 0.743 & 0.577 & 0.938 & 0.935 & 0.896\\

HEAL (L) \cite{c2} & 0.832 & 0.790 & 0.623 & 0.968 & 0.963 & 0.926\\
\hline
\rowcolor{blue!20}
DRCP (LC) & \textbf{0.878} & \textbf{0.836} & \textbf{0.674} & \textbf{0.978} & \textbf{0.976} & \textbf{0.951}\\
\hline 
\end{tabular}%
}
\caption{Comparison of existing cooperative perception methods with our proposed DRCP across different datasets. Each baseline is evaluated under its best-performing modality configuration, as reported in the original papers and verified in our experiments (LC denotes LiDAR–camera, and L denotes LiDAR-only). Notably, DRCP is the only diffusion-based approach in the comparison.}
\label{tab:perf}
\end{table}

\begin{figure*}[htbp]
    \vspace{2mm}
    \centering
    \begin{subfigure}[b]{0.329\textwidth}
        \centering
        \includegraphics[width=\textwidth]{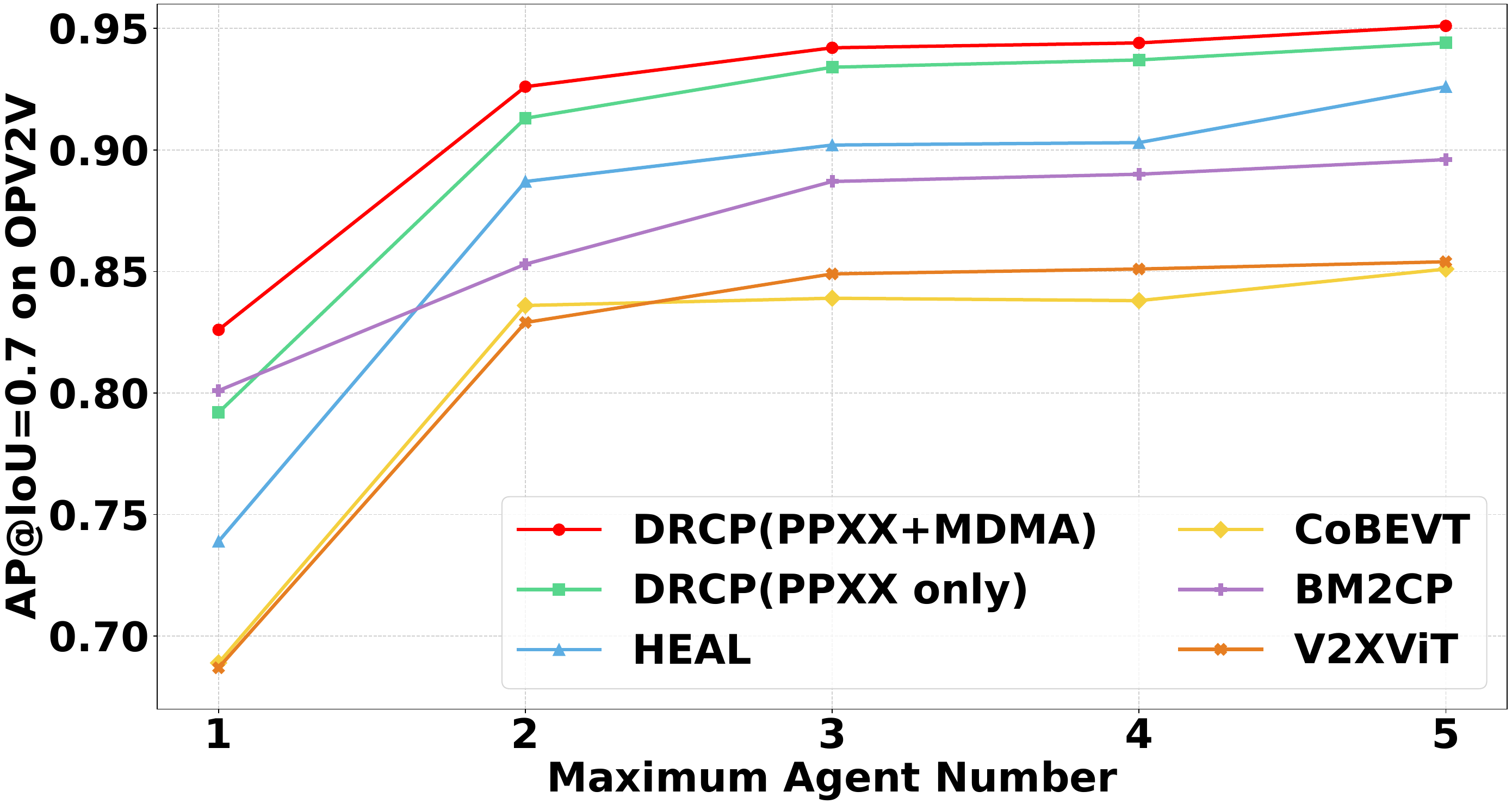}
        \caption{AP@IoU=0.7 across agent numbers}
        \label{fig:sub1}
    \end{subfigure}
    \hfill
    \begin{subfigure}[b]{0.329\textwidth}
        \centering
        \includegraphics[width=\textwidth]{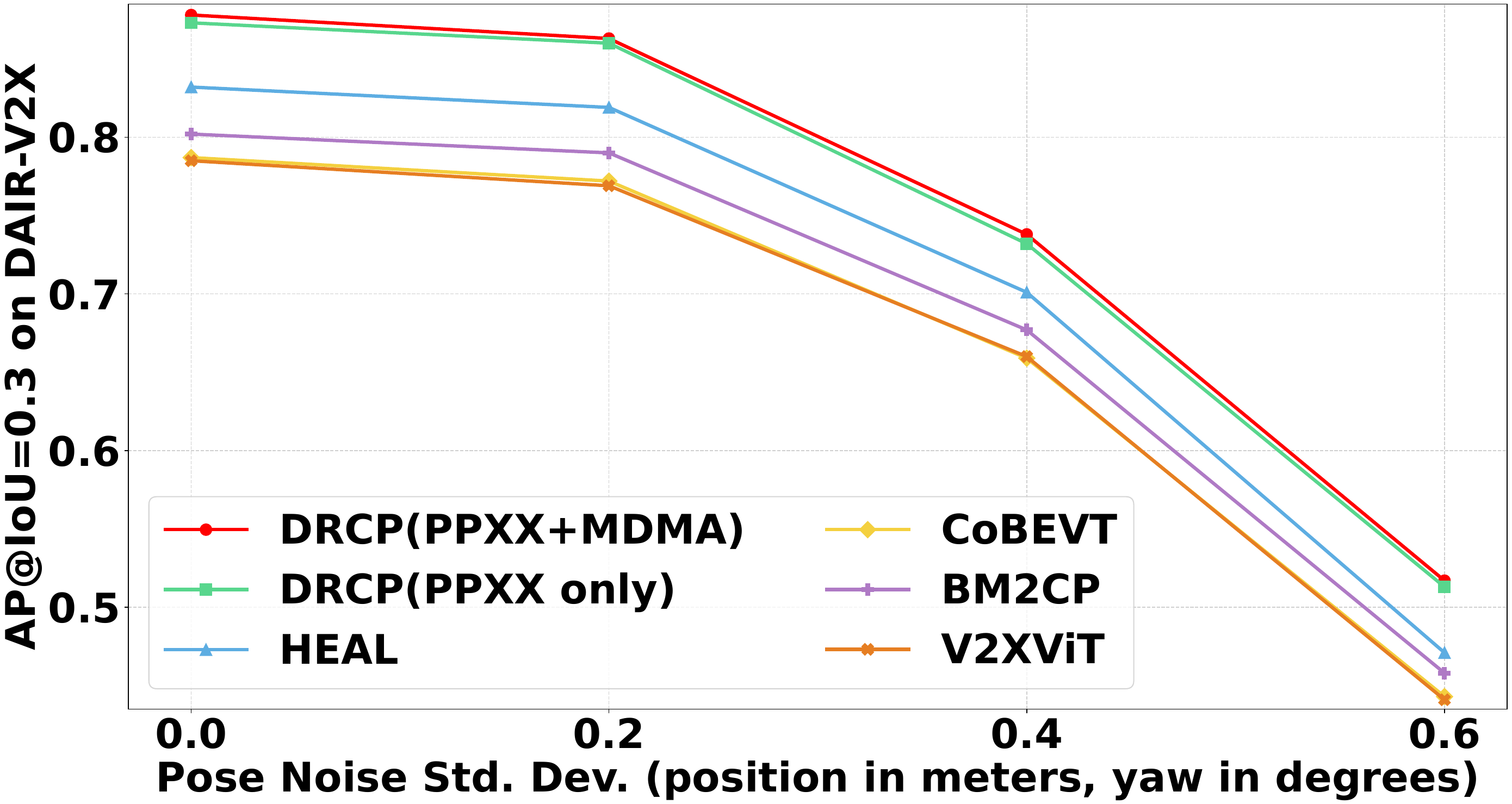}
        \caption{AP@IoU=0.3 across pose noises}
        \label{fig:sub2}
    \end{subfigure}
    \hfill
    \begin{subfigure}[b]{0.329\textwidth}
        \centering
        \includegraphics[width=\textwidth]{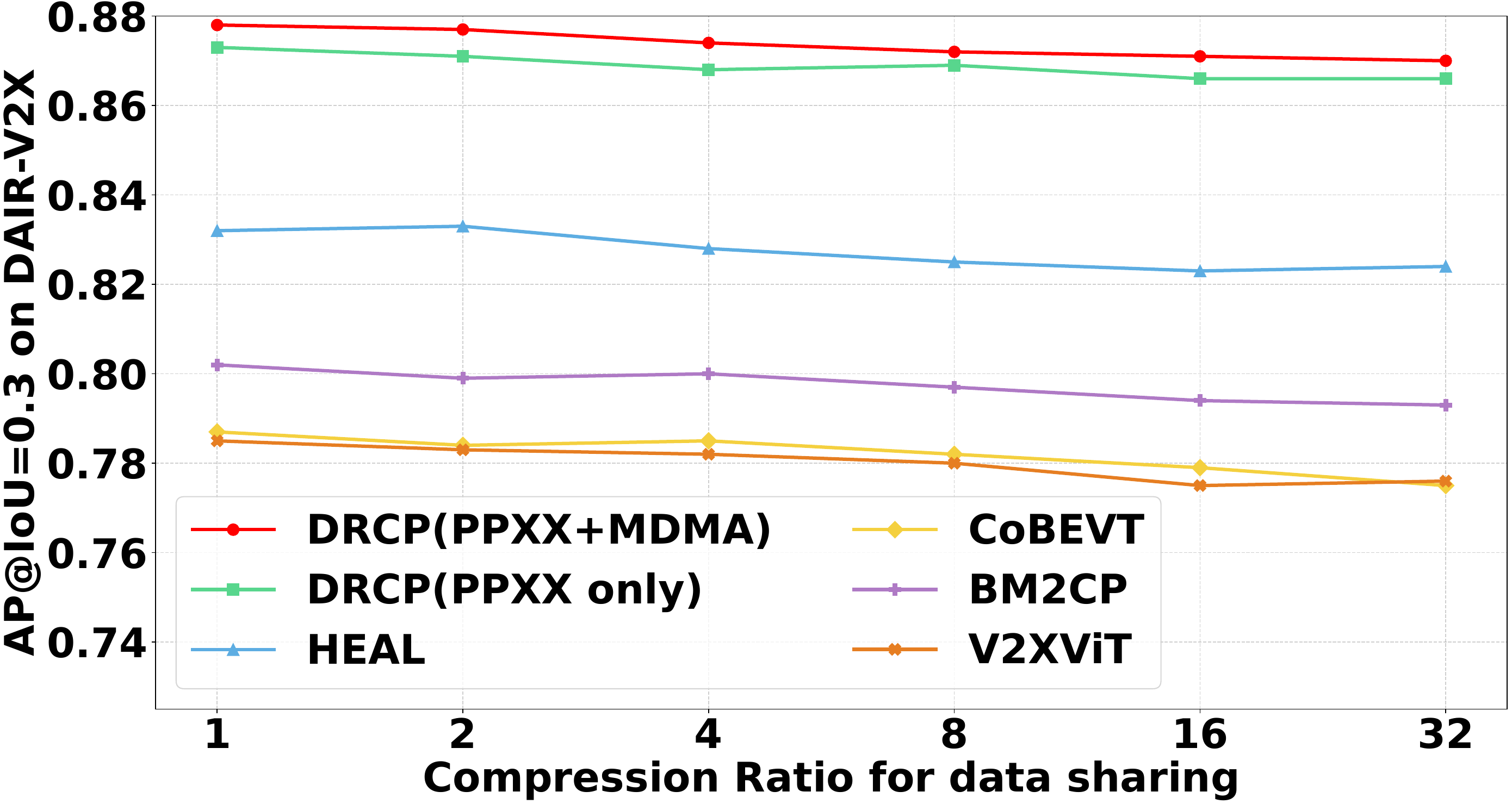}
        \caption{AP@IoU=0.3 across compression ratios}
        \label{fig:sub3}
    \end{subfigure}
    
    \caption{Object detection performances at: (a) varying maximum numbers of collaborative agents (ground truth assigned based on the actual participating agents.), (b) varying levels of pose noises and (c) varying ratios of sharing data compression.}
    \label{fig:ap30s}
    \vspace{-2mm}
\end{figure*}

To further demonstrate our method’s effectiveness—particularly the MDMA module—we evaluate how the number of participating agents affects perception (\cref{fig:sub1}). With two cooperating agents, OPV2V AP70 improves by 3.9\% over prior methods, while additional agents yield diminishing gains, indicating that fewer agents lead to noisier and less complete perception, providing greater scope for feature refinement. Detection ground truths are assigned only to participating agents, emphasizing enhancement of BEV representations via our modules rather than extended ego-vehicle coverage. In \cref{tab:cp_effective}, where ground truth is consistently based on two agents, comparing single-agent versus two-agent collaboration reveals at least a 10.0\% improvement across settings (PPXX only or PPXX+MDMA as DRCP), underscoring the substantial benefits and necessity of multi-agent cooperation.

An intriguing observation emerges from the OPV2V single-agent results (see \cref{tab:cp_effective} or \cref{fig:sub1}), where MDMA boosts detection performance by approximately 2.0\%. In contrast, DAIR-V2X shows a smaller but notable improvement. This discrepancy likely stems from factors such as the heterogeneous sensor configurations in DAIR-V2X versus the uniform setup in OPV2V, or the greater complexity of real-world conditions. Further studies on MDMA’s effect in single-agent perception (e.g., on NuScenes) are therefore warranted before drawing definitive conclusions.

Moreover, we evaluated robustness to pose noise, as illustrated in \cref{fig:sub2}, by adding Gaussian perturbations at varying levels to the original pose data in DAIR-V2X. Our method consistently maintains its performance advantage across all noise levels, demonstrating robustness to localization noise and stability of the fusion mechanisms.

\begin{table}[h]
\centering
\resizebox{\columnwidth}{!}{%
\begin{tabular}{c|c|c|c|c|c|c|c} 
\hline
\multicolumn{2}{c|} {Dataset} & \multicolumn{3}{c|}{DAIR-V2X} & \multicolumn{3}{c}{OPV2V} \\
\hline
Method & Num & AP30 & AP50 & AP70 & AP30 & AP50 &AP70\\
\hline
PPXX & 1 & 0.720 & 0.683 & 0.570 & 0.840 & 0.827 & 0.722\\

PPXX & 2 & 0.873 & 0.831 & 0.668 & 0.963 & 0.959 & 0.913\\
\hline
PPXX+MDMA & 1 & 0.724 & 0.686 & 0.576 & 0.859 & 0.843 & 0.749\\

PPXX+MDMA & 2 & 0.878 & 0.836 & 0.674 & 0.971 & 0.966 & 0.926\\
\hline 
\end{tabular}%
}
\caption{Comparison of no collaboration versus collaboration, Num stands for actual participating number of agents. PPXX as Non-diffusion DRCP and PPXX+MDMA as Diffusion-enabled.}
\label{tab:cp_effective}
\end{table}

\noindent \textbf{Ablation Study:}
\cref{tab:ablation} presents a systematic evaluation of DRCP components and sub-modules on DAIR-V2X. The baseline—a basic cross-modal fusion module with intrinsics-agnostic radian-glue attention (evenly divided radian version) and Integrated Pyramid Fusion—already surpasses HEAL \cite{c2} by 2.8\%, 2.5\%, and 1.4\% in AP30, AP50, and AP70, highlighting the benefits of cross-modal fusion. Each added component further improves performance. Within PPXX, performance steadily increases as key sub-modules are incorporated. Intrin-RG-Attn notably enhances fine-grained spatial alignment, yielding a 1.5\% gain in AP70 versus 0.7\% in AP30. Adaptive Convolution at the final BEV stage exploits inter-agent dependencies across scales, enriching the BEV representation with collaborative cues and producing a more coherent, semantically informative embedding.

\begin{table}[h]
\centering
\small
{%
\begin{tabular}{c|c|c|c} 
\hline
Dataset & \multicolumn{3}{c}{DAIR-V2X} \\
\hline
Component & AP30 & AP50 & AP70 \\
\hline
Baseline & 0.860 & 0.815 & 0.637 \\
\rowcolor{yellow!10}
PPXX (Intrin-RG-Attn) & 0.867 & 0.822 & 0.652 \\
\rowcolor{yellow!10}
PPXX (Adaptive Convolution) & 0.869 & 0.824 & 0.651 \\
\rowcolor{yellow!30}
PPXX & 0.873 & 0.831 & 0.668 \\
MDMA (no mask)& 0.694 & 0.609 & 0.215 \\
\rowcolor{yellow!10}
MDMA (mask 1 only)& 0.833 & 0.747 & 0.407 \\
\rowcolor{yellow!10}
MDMA (mask 2 only)& 0.867 & 0.821 & 0.642 \\
\rowcolor{yellow!30}
MDMA & \textbf{0.879} & 0.833 & 0.658 \\
\rowcolor{yellow!60}
PPXX+MDMA & \textbf{0.878} & \textbf{0.836} & \textbf{0.674} \\
\hline 
\end{tabular}%
}
\caption{Component Ablation Study Comparison. The baseline is configured with the intrinsics-agnostic Radian-Glue Attention for cross-modal fusion and Integrated Pyramid Fusion for cross-modal fusion. PPXX and MDMA without parentheses represents all sub-components enabled.}
\label{tab:ablation}
\end{table}

\begin{table}[h]
\centering
\resizebox{\columnwidth}{!}{%
\begin{tabular}{c|c|c|c|c|c} 
\hline
Method & DRCP & HEAL & CoBEVT & BM2CP & V2X-ViT\\
\hline
Parameters & 44.1/38.5 M & 33.5 M& 44.2 M& 31.7 M& 49.8 M\\

Inference time & 68/54 ms & 43 ms& 179 ms& 57 ms& 142 ms\\
\hline 
\end{tabular}%
}
\caption{Comparison of model size (millions of parameters) and inference speed (ms) for different methods; DRCP shows two configurations: PPXX+MDMA/PPXX-only.}
\label{tab:inference_speed}
\end{table}

The MDMA diffusion module further refines BEV features by injecting structured, learnable uncertainty, extrapolating information beyond perceived features into a less ambiguous, task-optimal manifold. Standalone, it yields 1.9\%, 1.6\%, and 2.1\% gains on AP30, AP50, and AP70. While overlap with PPXX reduces incremental gains, MDMA complements PPXX by providing a cleaner, more task-aligned BEV representation. This synergy is more evident under reduced-agent settings: on OPV2V, AP70 drops by 2.7\% with one agent and 1.3\% with two agents when MDMA is removed (\cref{fig:sub1}, \cref{tab:cp_effective}). Ablations further confirm the necessity of the diffusion masks: removing both causes severe degradation, as the unconditioned one-step diffusion cannot reconstruct the BEV; using only the first mask underperforms the no-diffusion baseline, since it solely guides the denoiser toward reconstructing the BEV; the second mask alone offers limited benefit, preserving strong original features and introducing positiveness. The full MDMA setup with both masks optimally fuses conditioned one-step denoising with the original BEV, enabling numerically stable refinements that enhance semantic alignment and perceptual salience, thus validating the proposed masking strategy.

\noindent \textbf{Computation \& Communication Budget:}
The proposed DRCP framework achieves cooperative perception at 68 ms per frame (from raw data to final BEV) under a 4 MB sensory-sharing bandwidth. The most lightweight configuration---PPXX alone---runs at 54 ms per frame, already surpassing prior methods and supporting real-time inference, leaving headroom for downstream planning and control. Component-wise, Intrin-RG-Attn runs in $\sim$9.5 ms, Adaptive Convolution adds $<$1 ms, and the MDMA diffusion module requires $\sim$15 ms, keeping the pipeline feasible for latency-sensitive deployment. While a 4 MB budget may appear prohibitive for real-world V2X, a lightweight autoencoder achieves up to 32$\times$ compression (0.125 MB) with only 0.8\% accuracy loss and an additional 2 ms runtime (\cref{fig:sub3}). Besides, we also compare model size and inference speed with key baselines in \cref{tab:inference_speed}.

\noindent \textbf{Other Settings:}  
We also tested Swin Transformer blocks \cite{Swin} with different window sizes as alternatives to Adaptive Convolution. None succeeded; even the best variant (window=8) suffered AP drops of 0.4\%, 0.7\%, and 3.7\% on DAIR-V2X, confirming the effectiveness of Adaptive Convolution.  

To evaluate the scheduler effect in MDMA, we compared DDIM and DDPM under varying timesteps. Unlike conventional diffusion settings, our framework treats the scheduler primarily as a stochastic noise injector for generating diverse feature candidates, followed by one-step refinement. Both methods showed comparable performance between 5 and 20 steps, with DDIM reaching its best at 15 steps and DDPM peaking at 20 steps. This suggests that DDIM can more quickly provide sufficient noise diversity, while DDPM, being closer to pure Gaussian injection, yields stronger results when given more steps (we adopt 20 steps DDPM).

Moreover, applying MDMA in a LiDAR-only BEV setting yielded moderate yet consistent gains (e.g., +0.7\% in AP70 on DAIR-V2X), underscoring the role of camera-derived semantics in guiding diffusion and refining spatial features.

\begin{figure}[t]
  \vspace{2mm}
  \centering
  \begin{subfigure}{0.493\linewidth}
    \centering
    \includegraphics[width=1.0\linewidth]{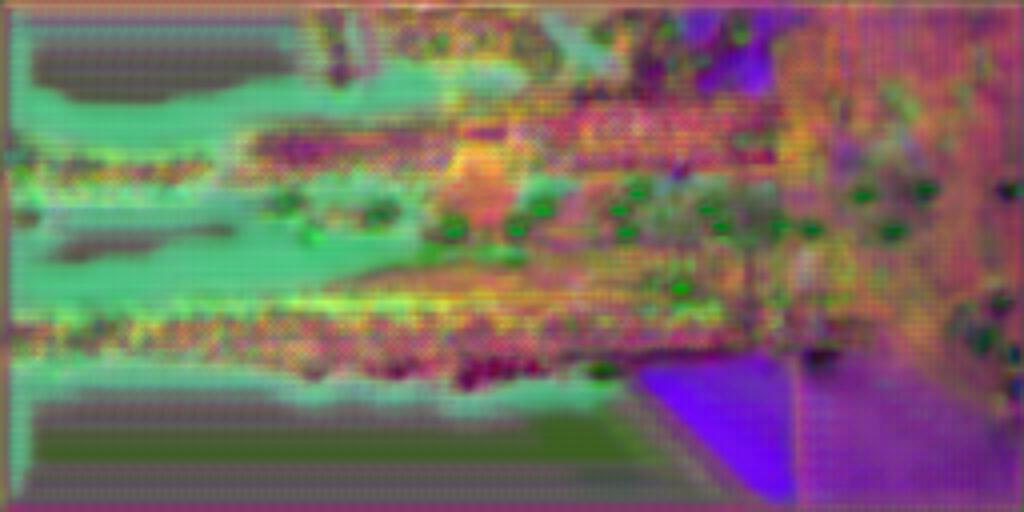}
    \caption{Original BEV}
    \label{Fig:v1}
  \end{subfigure}
  \hfill
  \begin{subfigure}{0.493\linewidth}
    \centering
    \includegraphics[width=1.0\linewidth]{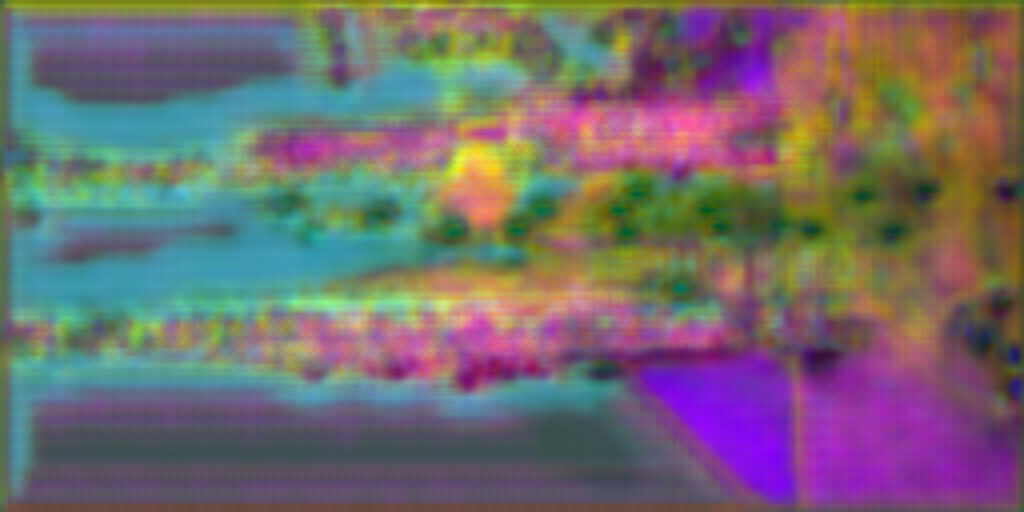}
    \caption{MDMA Diffused BEV}
    \label{Fig:v2}
  \end{subfigure}
  
  \vspace{0.5em}
  
  \begin{subfigure}{0.493\linewidth}
    \centering
    \includegraphics[width=1.0\linewidth]{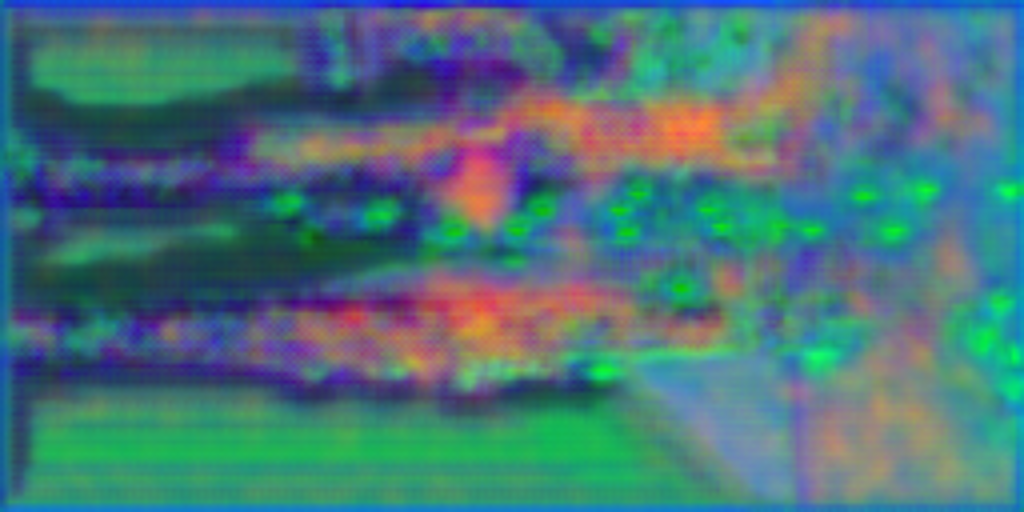}
    \caption{Residual BEV}
    \label{Fig:v3}
  \end{subfigure}
  \hfill
  \begin{subfigure}{0.493\linewidth}
    \centering
    \includegraphics[width=1.0\linewidth]{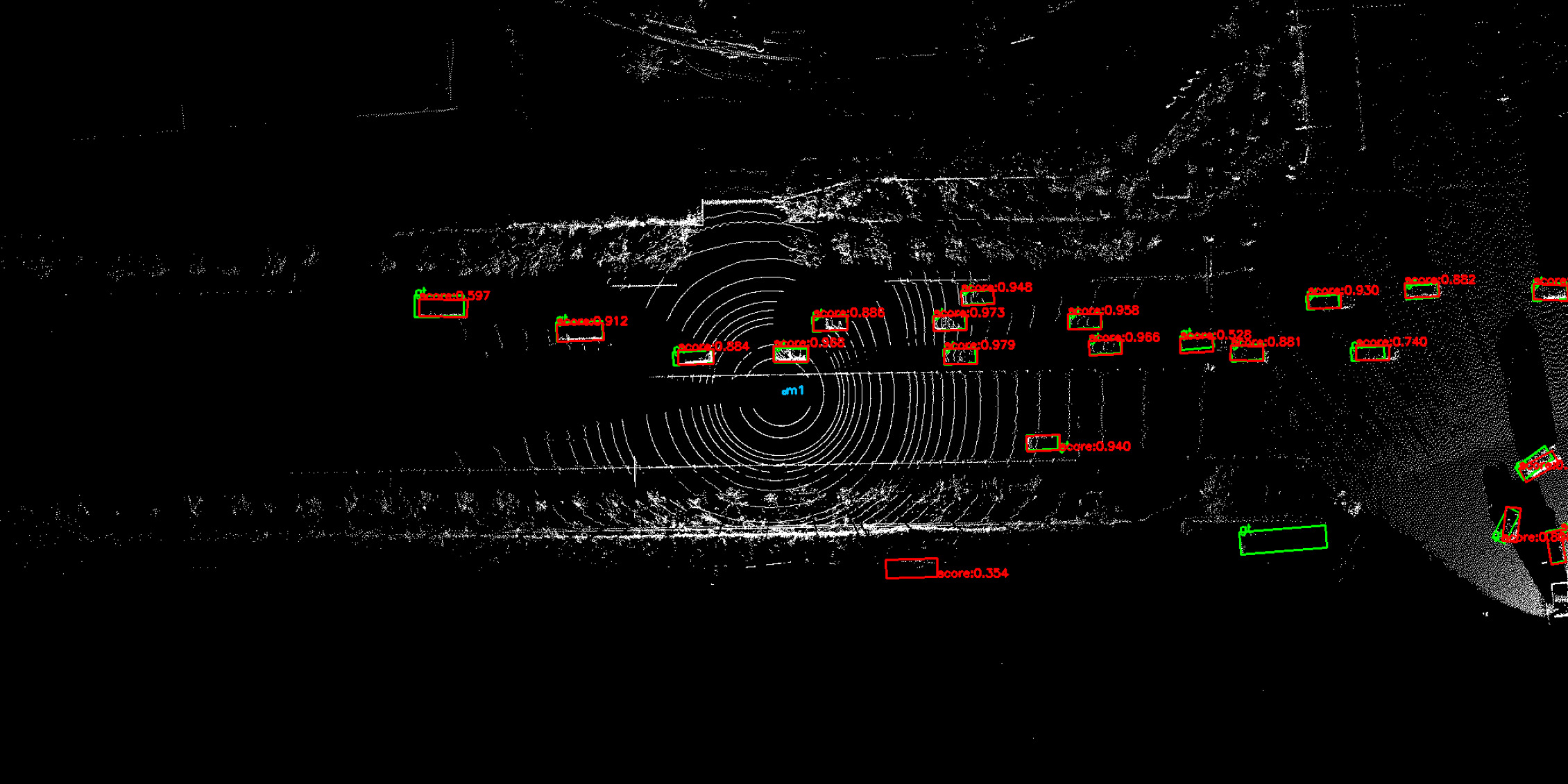}
    \caption{Detection Result}
    \label{Fig:v4}
  \end{subfigure}
  
  \caption{PCA visualizations of BEV features before (a) and after (b) MDMA refinement. Residual enhancements are highlighted in (c), while downstream detection results are shown in (d).}
  \label{fig:visualization}
  \vspace{-2mm}
\end{figure}

\noindent \textbf{Visualizations of BEV:}
The PCA-based visualizations in \cref{fig:visualization} offer a global perspective on how the MDMA module refines BEV features. While the original (\cref{Fig:v1}) and enhanced maps (\cref{Fig:v2}) appear visually similar at first glance, the enhanced BEVs exhibit more spatially coherent clusters and improved semantic alignment, which is corroborated by detection outputs (\cref{Fig:v4}). The residual map (\cref{Fig:v3}), obtained by subtracting the original from the enhanced BEV, makes this effect explicit: target objects gain enlarged and better-contrasted footprints relative to the background.

\begin{figure}[!h]
  \vspace{2mm}
  \centering
   \includegraphics[width=1.0\linewidth]{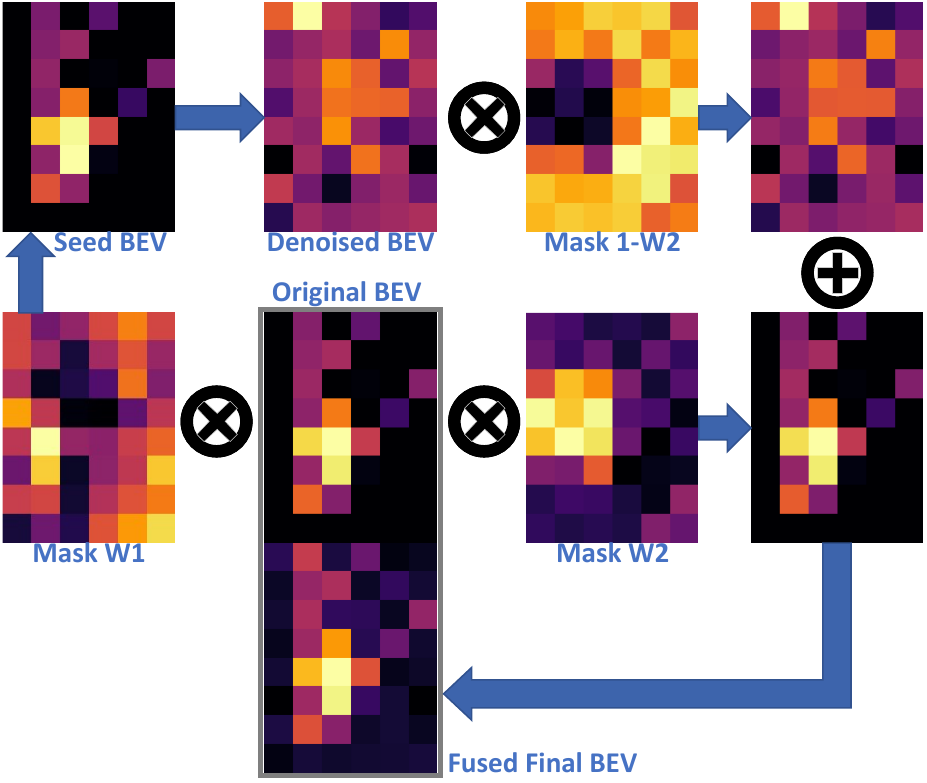}
   \caption{Representative channel-wise heatmaps across the MDMA pipeline. All visual differences are relative to each tensor’s own dynamic range.}
   \label{fig:mdma_ht}
\end{figure}

To examine finer-scale effects, we inspect localized heatmaps around challenging targets highlighted in \cref{fig:concept} as shown \cref{fig:mdma_ht}. Although diffused maps initially seem to introduce only faint activations in previously blank regions, these weak signals are present per channel. Given that the BEV tensor contains hundreds of channels (i.e., 256), these small, distributed residuals accumulate into substantial cross-channel reinforcement, effectively shifting the manifold toward task-aligned representations. The second mask in particular introduces denoised compensation features in blank regions, contributing to a more complete representation.

Quantitatively, beyond the color-scale differences observed in the heatmaps above, the process can be further interpreted through explicit numerical analysis. The first mask compresses the dynamic range of the original BEV (i.e., $[0.0,0.68] \rightarrow [0.0,0.33]$), keeping the seed in a numerically stable regime and preventing extreme one-step fluctuations. The denoiser outputs modest corrective signals (i.e., [-0.06,0.07]), whose weighted residuals lie approximately in $[-0.03,0.04]$. After fusion, the final BEV spans roughly $[-0.03,0.37]$, indicating that MDMA operates in a high Signal-to-Noise-Ratio (SNR), low-variance regime: rather than “redrawing” features, the single-step pass performs gentle residual correction. From a signal-processing perspective, this aligns the conditioning SNR with the denoiser’s capacity for stable guidance. From a learning standpoint, the aggregated multi-channel residuals strengthen perceptual salience and semantic calibration of object features, directly supporting the improved detection performance observed.

\section{Conclusion}
We presented DRCP, a real-time cooperative perception framework that integrates a cross-modal, cross-agent fusion backbone with a lightweight, single-step diffusion module for BEV feature refinement. The framework adaptively enhances ambiguous features and aligns BEV representations to task-optimal manifolds, improving multi-agent perception performance. Future work will explore temporal modeling and dynamic information sharing to strengthen generalization and robustness in real-world deployments.

\bibliographystyle{IEEEtran}
\bibliography{main}

\end{document}